\documentclass[sigconf]{acmart}
\usepackage{subcaption}

\AtBeginDocument{%
  \providecommand\BibTeX{{%
    \normalfont B\kern-0.5em{\scshape i\kern-0.25em b}\kern-0.8em\TeX}}}

\acmConference[UrbComp Workshop '21]{UrbComp Workshop '21: ACM SIGSPATIAL}{November 2021}{Beijing, China}

\begin{document}

\title{Attention-based Contextual Multi-View Graph Convolutional Networks for Short-term Population Prediction}


\author{Yuki Kubota}
\affiliation{%
  \institution{Tokyo Metropolitan
  University}
  \city{Tokyo}
  \country{Japan}}
\email{yuki9802251597@gmail.com}


\author{Yuki Ohira}
\affiliation{%
  \institution{Tokyo Metropolitan
  University}
  \city{Tokyo}
  \country{Japan}}
\email{ohira@tmu.ac.jp}

\author{Tetsuo Shimizu}
\affiliation{%
  \institution{Tokyo Metropolitan
  University}
  \city{Tokyo}
  \country{Japan}}
\email{t-sim@tmu.ac.jp}

\renewcommand{\shortauthors}{Yuki Kubota, Yuki Ohira, Tetsuo Shimizu}

\begin{abstract}
  Short-term future population prediction is a crucial problem in urban computing. Accurate future population prediction can provide rich insights for urban planners or developers.
  However, predicting the future population is a challenging task due to its complex spatiotemporal dependencies. 
  Many existing works have attempted to capture spatial correlations by partitioning a city into grids and using Convolutional Neural Networks (CNN).
  However, CNN merely captures spatial correlations by using a rectangle filter; it ignores urban environmental information such as distribution of railroads and location of POI.
  Moreover, the importance of those kinds of information for population prediction differs in each region and is affected by contextual situations such as weather conditions and day of the week.
  To tackle this problem, we propose a novel deep learning model called
 Attention-based Contextual Multi-View Graph Convolutional Networks (ACMV-GCNs).
 We first construct multiple graphs based on urban environmental information, and then ACMV-GCNs captures spatial correlations from various views with graph convolutional networks.
Further, we add an attention module to consider the contextual situations when leveraging urban environmental information for future population prediction. 
  Using statistics population count data collected through mobile phones, we demonstrate that our proposed model outperforms baseline methods. In addition, by visualizing weights calculated by an attention module, we show that our model learns an efficient way to utilize urban environment information without any prior knowledge.
\end{abstract}

\ccsdesc[500]{Human-centered computing~Ubiquitous and mobile computing}
\ccsdesc[300]{Computing methodologies~Artificial intelligence}

\keywords{Spatio-temporal data, Neural Networks, Graph Convolutional Networks, Population Prediction}


\maketitle

\section{Introduction}
The recent development of positional measurement technology allows us to collect diverse data generated in urban areas. Analyzing such data will provide benefits for us to understand urban status quantitatively.  
In particular, predicting future population in a city has a 
great potential to provide rich insights for 
wide range of applications such as traffic management and public safety\cite{urban_computing}.
In this paper, we attempt to predict short-term future populations among all regions in a city by using historical population count data.
However, this is a very challenging problem since data collected in urban areas contain complex spatiotemporal dependencies
\cite{urban_flow_servey}.

Typically, the autoregressive integrated moving average (ARIMA) and its variants, which statistically handle time-series data, have been used for spatiotemporal prediction problems\cite{arima_taxi}. 
However, while these models can capture temporal correlations, they still fail to capture the spatial correlations among regions\cite{urban_flow_servey}.

Therefore, some researchers have leveraged machine learning algorithms, which have achieved great success in various fields, to apply them in spatiotemporal prediction problems\cite{dmvst, crowd_flow_correlation, DeepSTN, multi_task}. 
The two popular models mainly used for spatiotemporal data analysis are Long Short-Term Memory (LSTM) and Convolutional Neural Networks (CNN).
LSTM is mainly used for time-series data analysis since it can capture temporal correlations by propagating hidden layer information along a time axis.
Although CNN is originally a model to process images, it can also be applied for spatial data analysis. For example, several researchers have partitioned a city into grids and then used CNN-based models to capture spatial correlations\cite{dmvst,ConvLSTM,dnn,residual}. 

Recently, several works have demonstrated that these deep learning-based models outperform classic statistical models\cite{dnn,dmvst,multi_task}.
However, CNN merely captures spatial correlations using a rectangle filter; it only captures spatial correlations based on Euclidean distance.
On the other hand, spatial correlations in a city are not limited to Euclidean distance, but various correlations caused by the surrounding environment (e.g., distribution of railroads, location of POI) also exist. 

We propose a novel method called Attention-based Contextual Multi-View Graph Convolutional Networks (ACMV-GCNs) to consider these unique spatial correlations in a city.
We first construct multiple graphs based on urban environmental information and then apply Graph Convolutional Networks (GCNs) for each of these graphs. This procedure allows the model to capture different spatial correlations from multiple views. 

We construct three graphs: distance graph, POI graph, and transportation graph. In general, the importance of these graphs for prediction differs in each region and is affected by contextual situations such as weather conditions and day of the week. 
For example, we can naturally assume that information about transportation systems is more crucial to predict metropolitan areas' population than suburban areas' population.
Besides, since most commercial facilities are closed at midnight, the contribution of information related to POI should be different at different times.

To handle this problem, we introduce a contextual-based attention module\cite{context_attention} that enables the model to learn the importance of different spatial views for each region based on contextual information.
In addition to its effectiveness, the attention weights can help us interpret the model prediction result.
In this paper, we later visualize the attention weights to confirm which urban environmental information contributes to population prediction in each region. 
In addition, we further investigate how these weights will fluctuate when affected by contextual information such as weather conditions. 

Although some existing works reveal that urban environmental information is beneficial to improving spatiotemporal prediction accuracy\cite{context_attention,dmvst,multi_view_gcn_sub}, there is a lack of investigation of how these kinds of information contribute to the prediction under various contextual situations.
Our proposed model could answer this question; to the best of our knowledge, it is the first such attempt in this field.
 
We summarize our contributions as follows:

\begin{itemize}
  \item We propose a novel method that captures unique spatial correlations in a city by efficiently using multiple graphs constructed based on different urban environmental information. 
  \item We evaluate our approach by using population count data collected through mobile phones and demonstrate the effectiveness of our framework.
  \item We investigate how urban environmental information contributes to population prediction and changes in different contextual situations by visualizing attention weights.
\end{itemize}

\section{Related work}
\label{sec:related_work}
Population prediction problems can be generalized as spatiotemporal data prediction problems.
Recently, several researchers have applied graph-based deep learning models for spatiotemporal data analysis.
One of the popular domains is traffic flow analysis. Since traffic sensors are fixed on a specific location, it is possible to construct graphs relatively easily. 
Indeed, a couple of works have leveraged GCNs for traffic data analysis and demonstrated its effectiveness\cite{traffic_attention,traffic_stgcn}.
However, these works defined the connectivity of each node based on Euclidean distance or past traffic flows pattern; that is, urban environmental information was not considered.

Some researchers leveraged urban environmental information to construct graphs, for example, by dividing the urban areas based on road networks\cite{multi_view_gcn_sub} or utilizing POI distribution\cite{context_attention}.
However, these works have constructed only one graph; they haven't constructed multiple graphs based on different views. Therefore, these works lack consideration for multiple spatial correlations.

Geng et al.\cite{ride_hailing_gcn} constructed multiple graphs from several aspects such as distribution of amusement facilities and connectivity of transportation networks for ride-hailing demand forecasting.
They predicted future demand for ride-hailing services with multi-graph convolutional networks and finally fused these output values.
Although the approach of this research is most related to our work, it still has several limitations.
First, this work didn't consider contextual information such as weather conditions and holiday information when fusing multiple output values. Instead, they proposed a method that combines multiple outputs by a simple aggregation function (e.g., sum, max, average, etc.)
Besides, this work only investigated the accuracy of a model; that is, it did not examine the contributions of each graph to prediction.

Our proposed method overcomes these limitations by incorporating the context-aware attention module that can assign different weights when fuses multiple output values by considering contextual information.
Therefore, our method can predict the future population more realistically. 
Besides, attention weights provide us a great insight into how contributions of each graph differ among regions and how they change in different contextual situations.
\section{Methodology}
\subsection{Problem Formulation}

In this section, we first define the future population prediction problem. We partition a city into an $ I \times J$ grid and represent an arbitrary grid as $v_n \in V$.
Here $I$ and $J$ is the number of rows and columns of the partitioned city, respectively, 
and $V$ represents the set of all regions.
Let $x_t^{n} \in \mathbb{R}$ be the population figure in a grid $v_n$ at time interval $t=\{1,\cdots,T\}$. We represent the population in entire urban regions at the $t$-th time interval as a tensor $\mathbf{X}_{t}\in \mathbb{R}^{1\times I\times J}$.

Hence, the future population prediction problem is formulated as learning the function that takes historical observations  $\{\mathbf{X}_{t-l}~ , \cdots , \mathbf{X}_{t}\} $ as input and predicts  $\mathbf{X}_{t+1}$. Here, $l$ denotes the length of the time interval that the model takes as input.

\subsection{Proposed Method}
\label{sec:architecture}
\begin{figure*}[h]
  \centering
  \includegraphics[width=15cm]{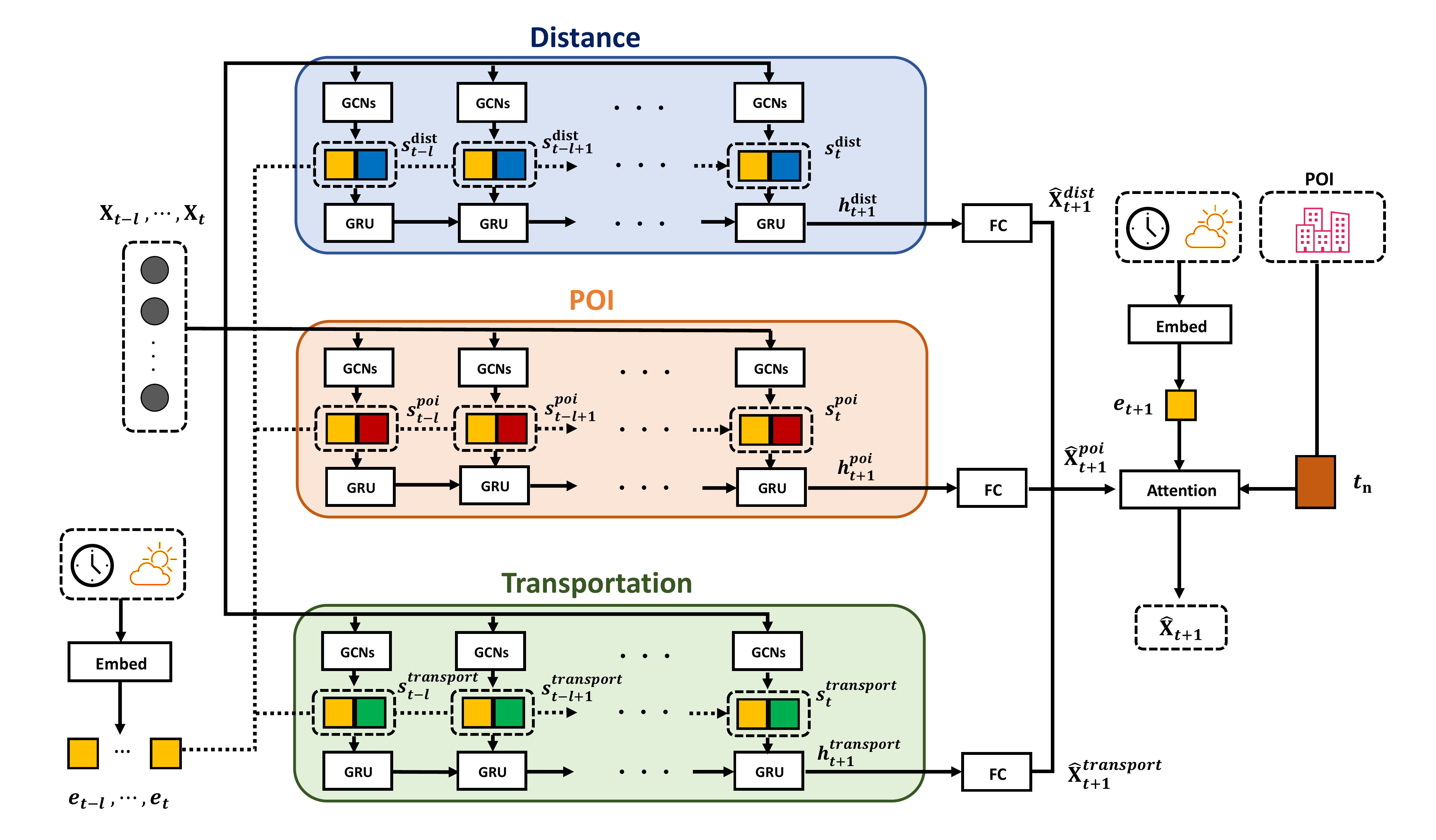}
  \caption{The architecture of the proposed ACMV-GCNs. We construct three parallel blocks to extract different aspects of relationships among regions, including distance, POI similarity, and connectivity of public transportation, by using multiple graphs. Each block contains GCNs and GRU to extract spatial and temporal patterns. The model ultimately fuses the resulting parallel hidden states by using an attention module, which considers contextual situations, to predict the future population.}
  \label{fig:architecture}
\end{figure*}

In this section, we provide details of our proposed method ACMV-GCNs, which contains both Graph Convolutional Networks (GCNs) and a contextual-based attention module.
Figure \ref{fig:architecture} shows the architecture of ACMV-GCNs.

The model contains three blocks in parallel to consider multiple spatiotemporal dependencies. Each block includes modules to capture spatial correlations and temporal correlations.
The model makes the final prediction by fusing three output values with the attention mechanism by considering contextual information.
Note that Embed and FC in Figure \ref{fig:architecture} represent an embedding layer and a fully connected layer, respectively.

\subsection{Construction of Graphs}
\label{sec:make_graph}
In this section, we provide details about how to construct multiple graphs based on urban environmental information.
We assumed that the graphs used in this paper are undirected.
A general formulation of a graph is represented as $\mathcal{G} = (V, \boldsymbol{A})$, 
where $V$ is a set of all regions in a city and $\boldsymbol{A} \in \mathbb{R}^{|V| \times |V|}$ is an adjacent matrix whose entries represent the strength of connectivity between two regions.

\subsubsection{Distance Graph}
\label{sec:dist_graph}
We first consider the Euclidean distance as one of the spatial correlations among a city.
We define the distance graph as $\mathcal{G}_{dist} = (V, \boldsymbol{A}_{dist})$, where $\boldsymbol{A}_{dist}$ denotes the adjacent matrix of the distance graph.
Each element of $\boldsymbol{A}_{dist}$ is calculated via a thresholded Gaussian kernel\cite{graph_signal_emerging} and defined as:

\begin{equation}
  \label{eq:A_dist}
  \boldsymbol{A}_{dist,nm} = \begin{cases}
                          \exp\bigl(-\frac{[dist(v_n,v_m)]^2}{2\theta^2}\bigr) & \text{if} \;\; dist(v_n, v_m) \leq \kappa \\
                          0 & \text{otherwise}
                         \end{cases}
\end{equation}
where $dist(n,m)$ represents the Euclidean distance between grid $v_{n}$ and $v_{m}$.
Let $\theta$ and $\kappa$ be two hyperparameters of the Gaussian kernel.

\subsubsection{POI Graph}
\label{sec:poi_graph}
POI is a specific point or location that attracts people's interest, such as a restaurant, park, and amusement facility.
Thus, we can assume that the POI distributions imply a characteristic of a region.
Since two regions with similar characteristics are highly related, it is natural to assume that we can exploit the spatial correlation from the similarity of POI distributions.
First, to obtain a region characteristics vector, we apply a TF-IDF model, which is frequently used in the natural language processing field.

We define the set of categories of POI as $C$ and the number of POIs located in the grid $v_n$ as $\{\omega_{n}^{c}, \forall c \in C\}$. 
Here $c$ denotes the specific category of POI.
Then, the value of TF-IDF of POI $c \in C$ in the grid $v_n$ is defined as:

\begin{equation}
  \label{eq:tf-idf}
  t_{n}^{c} = \frac{\omega_{n}^{c}}{\sum_{c \in C} \omega_{n}^{c}} \times \log\frac{|V|}{\bigl|\{v_{n} \in V|\omega_{n}^{c} \neq 0\}\bigr|}
\end{equation}
Here $|\cdot|$ denotes the cardinality of the set.
Applying Equation \ref{eq:tf-idf} to all categories of POI, we obtain a vector of POI distributions in grid $v_n$, which is denoted as 
$\boldsymbol{t}_{n} \in \mathbb{R}^{|C|}$.

Then, we calculate the cosine similarity between two vectors
and define an adjacency matrix of the POI graph as:
\begin{equation}
  \label{eq:A_poi}
  \boldsymbol{A}_{poi,nm} = \begin{cases} \cos(\boldsymbol{t}_{n}, \boldsymbol{t}_m) & \text{if} \;\; \cos(\boldsymbol{t}_{n}, \boldsymbol{t}_m)\geq \gamma \\
    0 & \text{otherwise},
  \end{cases}
\end{equation}
where $\gamma$ represents the threshold of similarity to control the sparsity of the adjacency matrix.
Finally, we represent the POI graph as
$\mathcal{G}_{poi} = (V, \boldsymbol{A}_{poi})$.

\subsubsection{Transportation Graph}
\label{sec:transportation_graph}
These days, most people who live in urban areas use transportation to move around a city. Thus, two regions sharing the same transportation networks can be correlated even if they are geometrically distant.
We consider a railroad network and a highway network as transportation networks. In particular, since people tend to gather at train stations, we connect two regions when they share stations that belong to the same railroad. In contrast, we make a connection for all regions when the same highway runs through them.

Let $M$ be the total category number of railroads and highways, where we distinguish each railroad and highways by its names, such as Yamanote-Line, Chyuou-Line, and Tomei Expressway.
We represent the distribution of transportation networks in grid $v_n$ as a vector $\boldsymbol{g}_{n} \in \mathbb{R}^{M}$.
We assign $1$ to each entry of the vector when the grid overlap with corresponding train stations or highway roads. Otherwise, the entry is set as $0$.
We take the inner product of two transportation's distribution vectors
and define an adjacency matrix of the transportation graph as:
\begin{equation}
  \label{eq:A_transport}
  \boldsymbol{A}_{transport,nm} = \boldsymbol{g}_{n}^\mathrm{T}\boldsymbol{g}_{m}.
\end{equation}
Note that the higher weight will be assigned to the entry when two regions share multiple transportation systems.
Finally, we define the transportation graph as $\mathcal{G}_{transport} = (V, \boldsymbol{A}_{transport
})$.

\subsection{Graph Convolutional Networks}
\label{sec:gcns}
Recently, applying neural network models to graphs has become an area of interest.
Deep learning methods for graphs can be grouped into two categories: spectral-based and spatial-based\cite{gcn_servey}. In this paper, we apply a spectral-based approach.

In spectral graph analysis, the value of each node is recognized as a signal and will be projected to a spectral domain by Graph Fourier Transform.
Hence, the convolutional operation for graph signal $\boldsymbol{x} \in \mathbb{R}^{N}$ is defined as:
\begin{equation}
  \label{eq:graphconv}
    g_{\boldsymbol{\theta}}*_{\mathcal{G}}\boldsymbol{x}
    = g_{\boldsymbol{\theta}} (\boldsymbol{L})\boldsymbol{x}
    = g_{\boldsymbol{\theta}}(\boldsymbol{U}\boldsymbol{\Lambda}\boldsymbol{U}^{\mathrm{T}})\boldsymbol{x}
    =\boldsymbol{U}g_{\boldsymbol{\theta}}(\boldsymbol{\Lambda})\boldsymbol{U}^{\mathrm{T}}\boldsymbol{x},
\end{equation}
where $*_{\mathcal{G}}$ denotes a graph convolution operation. $g_{\boldsymbol{\theta}} = \mathrm{diag}(\boldsymbol{\theta})$ denotes the convolution filter, which is parameterized by
$\boldsymbol{\theta}\in\mathbb{R}^{N}$.


$\boldsymbol{U} = [\boldsymbol{u_0},\boldsymbol{u_1}, \cdots ,\boldsymbol{u_{N-1}}] \in \mathbb{R}^{N \times N}$
is Fourier basis, and
$\boldsymbol{\Lambda} = \mathrm{diag}(\lambda_0, \lambda_1, \cdots ,\lambda_{N-1}) \in \mathbb{R}^{N \times N}$
is a diagnoal matrix in
which entries are eigen vectors and eigen values of the Laplacian matrix
$\boldsymbol{L} = \boldsymbol{I}_{N} - \boldsymbol{D}^{-\frac{1}{2}}\boldsymbol{A}\boldsymbol{D}^{-\frac{1}{2}} \in \mathbb{R}^{N \times N}$, respectively.
Here 
$\boldsymbol{I}_{N}\in \mathbb{R}^{N \times N}$
denotes a unit matrix,
and
$\boldsymbol{D}\in \mathbb{R}^{N \times N}$ is a diagonal matrix of node degrees $\boldsymbol{D}_{ii}=\sum_j{\boldsymbol{A}_{ij}}$.

However, the eigendecomposition of the Laplacian matrix $\boldsymbol{L}$ is highly computationally expensive; thus, applying graph convolution to massive graph data is non-trivial. 
Defferrand et al.\cite{chebNet} proposed Chebnet to tackle this issue.
ChebNet has succeeded in reducing the computational costs by approximating the filter $g_{\boldsymbol{\theta}}$ by Chebyshev polynomials.
The convolution operation of ChebNet is defined as:
\begin{equation}
  \label{eq:chebnet}
  g_{\boldsymbol{\theta}}*_{\mathcal{G}}\boldsymbol{x} = g_{\boldsymbol{\theta}} (\boldsymbol{L})\boldsymbol{x} = \sum_{k=0}^{K-1}\boldsymbol{\theta}_{k}T_{k}(\boldsymbol{\tilde{L}})\boldsymbol{x}b
\end{equation}
where
$\boldsymbol{\tilde{L}} = 2\boldsymbol{L}/\lambda_{max} - \boldsymbol{I}_{N}$. 
The recursive definition of the Chebyshev polynomial is $T_k(x) = 2xT_{k-1}(x) - T_{k-2}(x)$, where $T_{1}(x) = x$, $T_{0}(x) = 1$.
Calculating the $K$ order polynomial function using the filter is equivalent to making a convolutional operation by exploiting each node's $K$-hop nodes' information\cite{chebNet}.

\subsection{ACMV-GCNs Framework}
In this section, we explain the detail of the framework of ACMV-GCNs.
As shown in Figure \ref{fig:architecture}, the proposed method has three blocks parallel for each of the three graphs (i.e., distance, POI, and transportation).
The model calculates three output values by considering different spatial correlations based on different graphs. 
Finally, the model predicts the future population by fusing these values with considering contextual information by the attention module.

\subsubsection{Spatiotemporal correlation modeling}
\label{sec:spatiotemporal}
First, the model adopts the convolution operation denoted in Equation \ref{eq:chebnet} with different graphs represented by $\mathcal{G}_{dist}$, 
$\mathcal{G}_{poi}$, and $\mathcal{G}_{transport}$ to exploit the spatial correlations from multiple views.
To capture temporal correlations in addition to spatial correlations, we need to connect the output values of GCNs into a model that captures temporal correlations.
To achieve this, we set multiple GCNs units in a sequence where its number is the same as the input length $l$.

Hence, the grahph convolution operation of ACMV-GCNs at $t$-th time interval is denoted as:
\begin{equation}
  \label{eq:spatial_layer}
  \boldsymbol{H}_{t}^{(v+1)} = f(\sum_{k=0}^{K-1}\theta_{k}T_{k}(\boldsymbol{\tilde{L}})\boldsymbol{H}_{t}^{(v)}),
\end{equation}
where $\boldsymbol{H}_{t}^{(v+1)} \in \mathbb{R}^{N \times F_{v+1}}$ 
and
$\boldsymbol{H}_{t}^{(v)} \in \mathbb{R}^{N \times F_{v}}$
represent input and output of layer $v$ at the $t$-th time interval, respectively, and $f(\cdot)$ denotes an activation function.
The final output after $V$ times convolution operation is obtained as 
$\boldsymbol{H}_{t}^{(V+1)} \in \mathbb{R}^{N \times F_{V+1}}$.
We further flatten the output value into a vector; thus we obtain
$\boldsymbol{s}_{t}^{a} \in \mathbb{R}^{NF_{V+1}}$,
where $a \in \{dist,~ poi,~ transport \}$
represents which graph is used to calculate corresponding values.
Therefore, the final outputs of $l$ sequential GCNs are represented as
$\boldsymbol{s}_{t-l}^{a}, \cdots ,\boldsymbol{s}_{t}^{a}$.

In addition to the spatial correlations, we further consider the time of day, weather conditions, and holiday information as contextual information. 
The time of day variable contains $24$ values from $0$ to $23$.
The weather condition variable contains three categories: (1) sunny, (2) cloudy, and (3) rainy. Holiday information is represented by a dummy variable where $1$ means the corresponding times interval is a holiday.
In this paper, we assume that the weather condition at the specific time interval can be shared among all of the regions because we treat the relatively confined area as the target of the investigation.
The detail of the investigated area will be shown in the experiment part.
Considering this situation, we can share all of the contextual variables among whole regions in a city at every time interval. 

We treat all of contextual information variables as categorical features and convert them into a dense vector by using an embedding layer of neural networks.
Let $\mathbf{e}_{t} \in \mathbb{R}^{m}$ be an embedded vector of contextual information at the $t$-th time interval. 
Then, we further concatenate this vector with outputs of GCNs layers; thus, we obtain:
\begin{equation}
  \mathbf{p}_{t}^{a} = \mathbf{s}_{t}^{a} \oplus \mathbf{e}_{t}.
\end{equation}
Hence, vectors
$\mathbf{p}_{t-l}^{a}, \cdots, \mathbf{p}_{t}^{a}$
contain both spatial correlation and contextual information.

For the temporal correlation modeling, we add GRU\cite{GRU} as the following layer of GCNs.
GRU is one of the derivative models of LSTM.
The main difference between the two models is GRU
integrates both an input gate and a forget gate in LSTM 
into a single update gate, which successfully makes the model much simpler.
Considering GRU will take $\boldsymbol{p}_{t}^{a}$ as inputs, the 
updating formulation for each $a \in \{dist,~ poi,~ transport\}$ at the $t$-th time interval is defined as follows:
\begin{gather}
  \boldsymbol{z}_{t}^{a} = \sigma(\boldsymbol{M}_{z}^{a}\boldsymbol{p}_t^{a} + \boldsymbol{O}_{z}^{a}\boldsymbol{h}_{t-1}^{a} + \boldsymbol{b}_{z}^{a}) \\
  \boldsymbol{r}_{t}^{a} = \sigma(\boldsymbol{M}_{r}^{a}\boldsymbol{p}_t^{a} + \boldsymbol{O}_{r}^{a}\boldsymbol{h}_{t-1}^{a} + \boldsymbol{b}_{r}^{a}) \\
  \boldsymbol{\tilde{h}}_{t}^{a} = \phi(\boldsymbol{M}_{h}^{a}\boldsymbol{p}_t^{a} + \boldsymbol{r}_{t}^{a} \odot \boldsymbol{O}_{h}^{a}\boldsymbol{h}_{t-1}^{a} + \boldsymbol{b}_{h}^{a})  \\
  \boldsymbol{h}_{t}^{a} = (\boldsymbol{1} - \boldsymbol{z}_{t}^{a}) \odot \boldsymbol{h}_{t-1}^{a} + \boldsymbol{z}_{t}^{a} \odot \boldsymbol{\tilde{h}}_{t}^{a},
\end{gather}
where
$\boldsymbol{M}_{\beta}^{a},\boldsymbol{O}_{\beta}^{a}, \boldsymbol{b}_{\beta}^{a}~ (\beta \in \{z,r,h\})$
are all learnable parameters.
We finally obtain
$\boldsymbol{h}_{t+1}^{a}$ 
as the output value, which consider both spatiotemporal correlation and contextual information.

\subsubsection{Multi-View Fusion by the Attention Mechanism}
\label{sec:fusion}
We apply one fully connected layer to the output value
$\boldsymbol{h}_{t+1}^{a}$
and reshape it into a tensor, which is denoted as 
$\mathbf{\hat{X}}_{t+1}^{a} \in \mathbb{R}^{1 \times I \times J}~ (a \in \{dist,~ poi,~ transport\})$.
Since we obtain three tensors calculated based on each of the graphs
$\mathcal{G}_{dist}$, $\mathcal{G}_{poi}$, and $\mathcal{G}_{transport}$,
it is necessary to
integrate these values to make final prediction.
As we mentioned earlier, in reality, the importance of these three aspects will differ in each region and be affected by contextual situations.
To consider this fact in our model, we use a context-aware attention module whose basic idea was proposed in a 
taxi demands prediction work\cite{context_attention}.

\begin{figure}[h]
  \centering
  \includegraphics[width=9.cm]{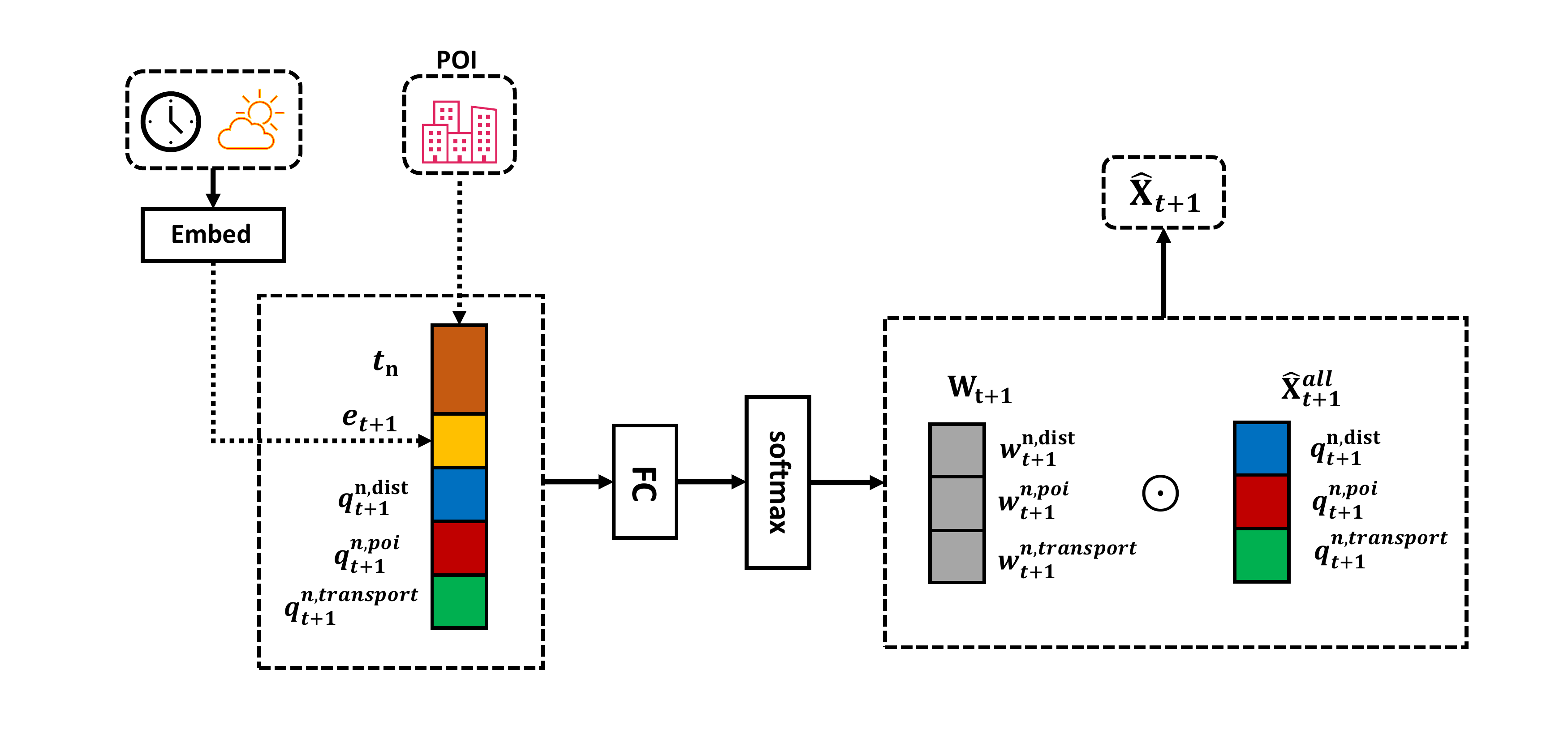}
  \caption{The architecture of a context-aware attention module used in ACMV-GCNs. The attention weights are calculated by considering contextual information for the predicting time interval and POI distribution of each region. Ultimately, the module fuses multiple predicted values by combining with calculated attention weights.}
  \label{fig:attention}
\end{figure}

Figure \ref{fig:attention} represents the architecture of the context-aware attention module used in ACMV-GCNs.
First, we embed contextual information at the predicting time interval $t+1$ by an embedding layer and obtain 
$\mathbf{e}_{t+1} \in \mathbb{R}^m$.
Let 
$q_{t+1}^{n,a} \in \mathbb{R}$
be a predicted value in grid
$v_n$ calculated based on graph $\mathcal{G}_a$.
For each grid, we combine these 
three predicted values
$q_{t+1}^{n,dist}$, $q_{t+1}^{n,poi}$, and $q_{t+1}^{n,transport}$
with 
a contextual information vector 
$\boldsymbol{e}_{t+1}$
and
the POI distribution vector
of each region denoted as
$\boldsymbol{t}_{n} \in \mathbb{R}^{|C|}$, 
which detail was introduced in Section 
\ref{sec:poi_graph}.
Note that although predicted values and a POI distribution vector differ in each region, we use the same context information $\boldsymbol{e}_{t+1}$ among all regions at the $t+1$-th time interval.
This is because as we mentioned in Section \ref{sec:spatiotemporal}, we share the contextual information 
at the certain time interval among whole regions in a city.
Therefore, we obtain a $3+m+|C|$ dimension vector for each region.

We further apply two layers (a fully connected layer with three-dimension output and a softmax layer) to the concatenated vector in each region.
Let $\mathbf{W}_{t+1} \in \mathbb{R}^{3 \times I \times J}$ be a tensor that gathered output values of this operation among all regions.
Hence, we can interpret that 
$\mathbf{W}_{t+1}$
is a weight tensor representing the importance for prediction of different spatial correlations aspects (i.e., distance, POI, and transportation) in each region at the time interval $t+1$.
Note that the calculation of these weights is accomplished by considering contextual information $\boldsymbol{e}_{t+1}$ and POI distribution $\boldsymbol{t}_{n}$.

After the calculation of the weight tensor $\mathbf{W}_{t+1}$,
we concatenate three predicted tensors
$\mathbf{\hat{X}}_{t+1}^{dist}$, 
$\mathbf{\hat{X}}_{t+1}^{poi}$, and 
$\mathbf{\hat{X}}_{t+1}^{transport}$
to obtain 
all predicted values among all regions, which is represented as a tensor 
$\mathbf{\hat{X}}_{t+1}^{all} \in \mathbb{R}^{3 \times I \times J}$.
Since both
$\mathbf{W}_{t+1}$
and
$\mathbf{\hat{X}}_{t+1}^{all}$
have an identical shape, we can take an element-wise product of these two tensors and sum up values for each region to finally obtain predicted values 
$\mathbf{\hat{X}}_{t+1} \in \mathbb{R}^{1 \times I \times J}$. 
We can interpret these procedures as taking the weighted average of three predicted values; those weights differ for each region and time interval.

Therefore, the predicted value for a region $v_n$ can be represented as:
  \begin{equation}
    \begin{split}      
      x_{t+1}^{n} = ~ &q_{t+1}^{n,dist}w^{n,dist}_{t+1}+q_{t+1}^{n,poi}w^{n,poi}_{t+1}\\
        &+ q_{t+1}^{n,transport}w^{n,transport}_{t+1},
    \end{split}
  \end{equation}
where
$w^{n,a}_{t+1}$ denotes the calculated weight for an output value of graph $\mathcal{G}_a$ in a grid $v_n$. 
Since we use a softmax layer to calculate these weights,
$w^{n,a}_{t+1} \geq 0$ and
$\sum_{a} w^{n,a}_{t+1}=1$ is established.

Note that if we aggregate multiple predicted values by simply taking the average, it is identical to assigning the  
same weight $w^{n,dist}_{t+1} = w^{n,poi}_{t+1} = w^{n,transport}_{t+1} = 1/3$ for all regions among all time intervals.
As we already mentioned, since the contributions of urban environmental information vary in different contextual situations, we can understand that this procedure does not reflect the real world's situation correctly.


\section{Experiment}
\subsection{Dataset Description}
In this paper, we use Mobile Spatial Statistics\footnote{https://mobaku.jp/}, massive population count data collected from mobile device's location information; provided by DOCOMO InsightMarketing, INC.
In provided data, the whole area of Japan is partitioned into $500m \times 500m$ size grids. 
The hourly population figures are estimated based on the logs of mobile devices and recorded for each grid. 
The dataset was collected from 1/1/2020 to 5/31/2020, and additional information about the people attribute, i.e., the age and the gender or residence (prefecture or municipality), is provided. 
Because of the need for anonymity of the dataset, 
these attributes were not provided altogether but separately.
Since the people who stay at home are also counted as population figures in this dataset, we extract the people who were visiting a municipality different from where they live and set it as the prediction target.

We obtain POI and transportation networks data in the Shapefile format from a website\footnote{https://nlftp.mlit.go.jp/ksj/} managed by the Ministry of Land, Infrastructure, Transport, and Tourism, a Japanese government ministry.
We separate POI into $9$ categories based on the information listed on the website.
We use open-source data provided by the Japan Meteorological Agency\footnote{https://www.jma.go.jp/jma/index.html} as weather information data.

\subsection{Experiment Settings}
\label{sec:setting}
\begin{figure}[h]
  \centering
  \includegraphics[width=7cm]{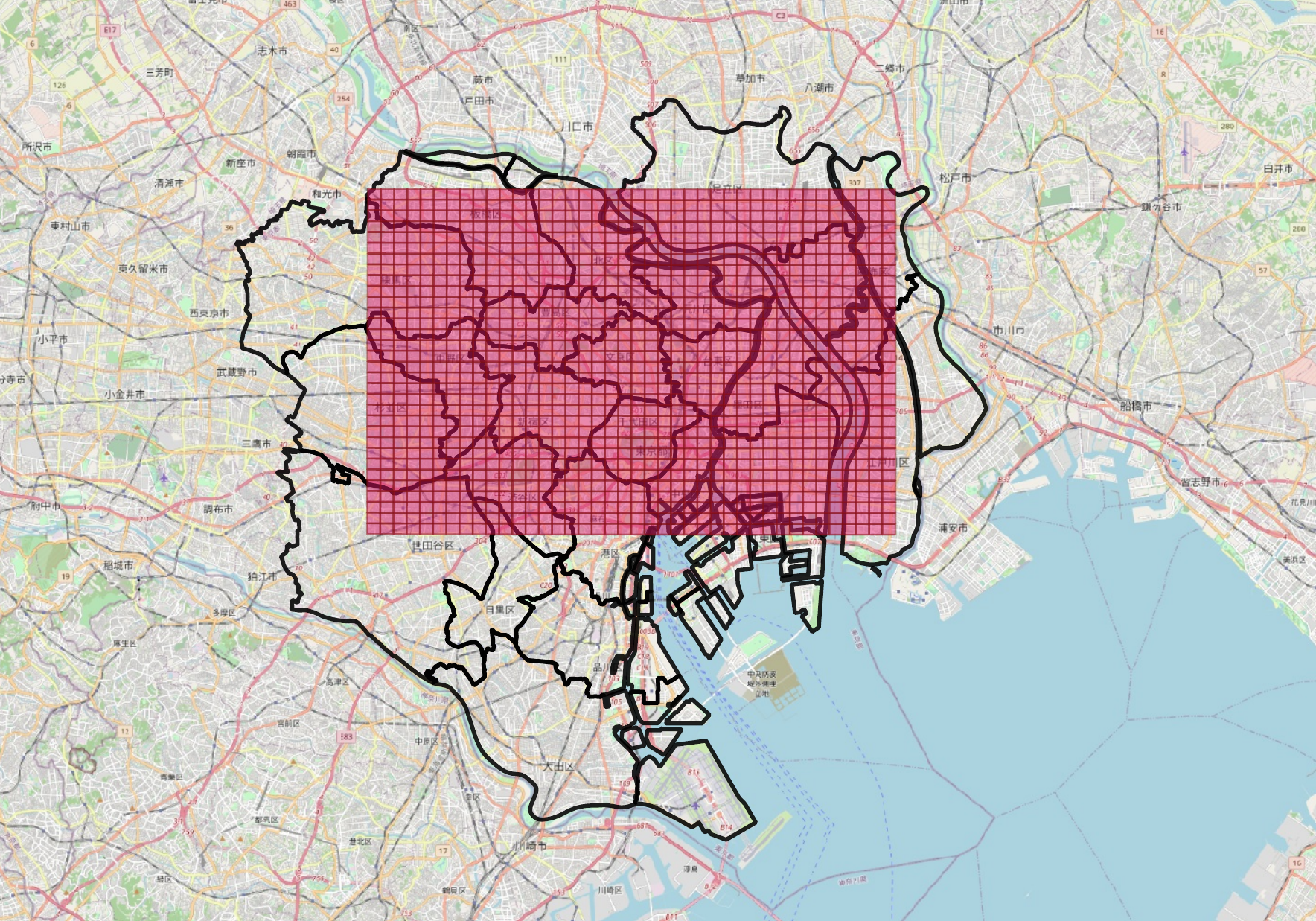}
  \caption{The investigated area in this paper is shown with red grids. Black lines indicate the boundaries of Tokyo 23 wards.}
  \label{fig:pred_area}
\end{figure}

In this paper, we select $32\times40$ size grids from the center of Tokyo as the investigated area.
The corresponding area is shown in Figure \ref{fig:pred_area}.

Considering the effect of COVID-19, we only use the dataset from 1/1/2020 to 3/24/2020. 
We then partition the dataset from 1/1/2020 to 3/10/2020 as train data, the set from 3/11/2020 to 3/17/2020 as validation data, and the set from 3/18/2020 to 3/24/2020 as test data.

To facilitate the training of neural network models, 
scaling the input values is known to be effective.
Considering the risk that the result of the scaling will be affected by extreme values, we use a scaling method known to be robust to the outlier, which is defined as:
\begin{equation}
  \hat{x_{i}} = \frac{x_{i} - Q_{1}}{Q_{3} - Q_{1}},
\end{equation}
where $Q_{1}$ and $Q_{3}$ are the $25th$ percentile and the $75th$ percentile of train data, respectively.
We later rescale the predicted values to obtain the actual population number and then evaluate the models' performance.

We use the optuna library\footnote{https://optuna.org/} for the hyperparameters tuning.
We choose the number of layers, the number of units, the activation function, and the learning rate for the target of searched hyperparameters.
We use Adam as the optimization algorithm.
The model is trained in the setting with the batch size of $32$ and the number of epochs as $200$.
We set the length of a sequence, which the model takes as input, as $8$, and the number of filters in GCNs as $K = 3$

\subsection{Evaluation Metric}
\label{sec:metrics}
We use Mean Absolute Error (MAE), Root Mean Square Error (RMSE), and Weighted Average Percentage Error (WAPE) for evaluation metrics, which are defined as follows:
\begin{gather}
  \textrm{MAE} = \frac{1}{NT} \sum_{n=1}^{N} \sum_{t=1}^{T} |y_{t}^{n} - \hat{y}^{n}_{t}| \\
  \textrm{RMSE} = \sqrt{\frac{1}{NT} \sum_{n=1}^{N}\sum_{t=1}^{T} (y_{t}^{n} - \hat{y}_{t}^{n})^{2}} \\
  \textrm{WAPE} = 100 \times \frac{1}{NT} \frac{\sum_{n=1}^{N}\sum_{t=1}^{T} |y_{t}^{n} - \hat{y}_{t}^{n}|}{\sum_{n=1}^{N}\sum_{t=1}^{T}|y_{t}^{n}|}
\end{gather}
where $y_{t}^{n}$ means the ground truth population number of grid $v_n$ at the $t$-th time interval and
$\hat{y}_{t}^{n}$ is the predicted value in the identical situation.
When evaluating the model performance, 
we trial training and prediction $10$ times identically for each model, 
then calculate the average score among those trials.

\subsection{Baselines}
\label{sec:baseline}
We compare the performance of ACMV-GCNs with the following methods, 
and tune the parameters for all methods in the same manner mentioned in Section \ref{sec:setting}.

\begin{itemize}
  \item \textbf{Historical Average (HA)}

  Historical average treats the population figure as a seasonal process. It predicts the population at certain time intervals according to the historical mean value of the same weekday and the same hour. 
  \item \textbf{Vector autoregression (VAR)}

  VAR is a model for forecasting time-series data by handling multiple time-series variables as a vector.
  \item \textbf{Convolutional Neural Network (CNN)}

  As already mentioned, CNN is a neural network model that captures spatial correlations; however, it doesn't consider the temporal correlations for prediction.
  \item \textbf{Long Short Term Memory (LSTM)}

  LSTM is also one of the neural network models.
  In contrast to CNN, LSTM predicts the population figure only based on temporal correlations.
  \item \textbf{Convolutional LSTM (ConvLSTM)}

  ConvLSTM\cite{ConvLSTM_origin} captures both spatial and temporal correlations simultaneously for prediction.

  \item \textbf{MV-GCNs(Multi-View Graph Convolutional Networks)}

  This model has an identical structure as ACMV-GCNs except for fusing three predicted values (i.e., distance-based, POI-based, and transportation-based) by simply taking their average. In contrast to our proposed method, which uses the attention module, this model assigns fixed weights for predicted values when fusing them regardless of the contextual situations.
\end{itemize}

\subsection{Results}
\label{sec:result}

\subsubsection{Performance comaprison}
Table \ref{table:result} shows the performance of the proposed method as compared to all other baseline models.
Note that the absolute value of the metrics is relatively high compared to existing works such as \cite{crowd_flow_correlation,crowd_flow_simple,multi_view_gcn_sub}.
This is because most of the existing works aim to predict the crowd flows, that is, the number of people who move to the other regions in the corresponding time intervals.
Besides, they usually use mobility trip data (e.g., TaxiNYC, BikeNYC, etc.) for 
the clowd flows prediction problems.
On the other hand, our research aims to predict the number of people who stayed in a certain region at the corresponding time intervals by directly using statistics population count data.
This difference causes that the values of the target of the prediction as well as metrics values in our research tend to show a higher value than the problems solved in existing works.

From Table \ref{table:result}, we can observe that ACMV-GCNs achieves the best performance in all metrics compared with other baseline models, which indicates the effectiveness of leveraging urban environmental information as well as contextual information for future population prediction.
Based on the fact that ConvLSTM achieves better performance than CNN and LSTM, we can validate the effectiveness of modeling spatial correlations and temporal correlations simultaneously.
Surprisingly, ConvLSTM outperforms MV-GCNs, which considers multiple spatial dependencies, in MAE and WAPE. 
One of the possible reasons could be some spatial aspects can deteriorate the model performance under specific contextual situations such as the transportation connectivity aspect at midnight.
Since MV-GCNs do not contain the attention module, it can not change the assigned weights even though the corresponding aspect is unimportant under the specific contextual situation.
Furthermore, we can observe that MV-GCNs performs worse than ACMV-GCNs in all metrics; this fact also indicates the effectiveness of fusing multiple predicted values by considering contextual situations with the attention module.

\begin{table}[h]
  \caption{Performance comaprison of different methods for population prediction}
  \label{table:result}
  \begin{center}
    \scalebox{1.0}[1.0]{
    \begin{tabular}{l|c|c|c}  \hline
      Model & MAE & RMSE & WAPE  \\ \hline
      HA & 205.64 & 980.68 & 19.20\% \\ 
      VAR & 91.30 & 165.93 & 18.94\% \\ 
      CNN & 65.27 & 121.25 & 17.45\% \\ 
      LSTM & 71.80 & 128.17 & 23.60\% \\  
      ConvLSTM & 59.85 & 111.55 & 15.27\% \\ 
      MV-GCNs & 61.03 & 109.25 & 18.88\% \\  \hline
      \textbf{ACMV-GCNs} & \textbf{57.00} & \textbf{108.73} & \textbf{14.11}\% \\ \hline
    \end{tabular}
    }
  \end{center}
\end{table}

\subsubsection{Performance on Different Graphs}
To investigate the effectiveness of spatial correlations modeling from 
multiple views, we evaluate the performance of the variants of ACMV-GCNs, which only use specific graphs among three of them.
The comparison result is shown in Table \ref{table:exclude}.
We can confirm that models based on a single graph perform worse than models based on multiple graphs. Further, ACMV-GCNs that uses all of the three graphs achieves the best performance.
Therefore, it is apparent that all three graphs used in this paper contribute to future population prediction. 
From this result,
we can confirm the effectiveness of combining multiple spatial correlations from different views.

\begin{table}[h]
  \caption{Comparison among models using different graphs}
  \label{table:exclude}
  \begin{center}
    \scalebox{1.0}{
    \begin{tabular}{l|c|c|c}  \hline
      Graph Used & MAE & RMSE & WAPE  \\ \hline
      Distance & 62.42 & 112.19 & 18.93\% \\ 
      POI & 65.65 & 121.31 & 19.81\% \\ 
      Transportation & 63.02 & 113.89 & 19.02\% \\  
      Distance + POI & 60.34 & 111.67 & 17.26\% \\ 
      Distance + Transportation & 58.86 & 108.82 & 15.95\% \\ 
      POI + Transportation & 60.26 & 112.82 & 16.32\% \\  \hline
      \textbf{ACMV-GCNs (All)} & \textbf{57.00} & \textbf{108.73} & \textbf{14.11}\% \\ \hline
    \end{tabular}
    }
  \end{center}
\end{table}

\subsubsection{Analysis of attention weights}
\begin{figure}[h]
  \centering
    \includegraphics[width=7cm]{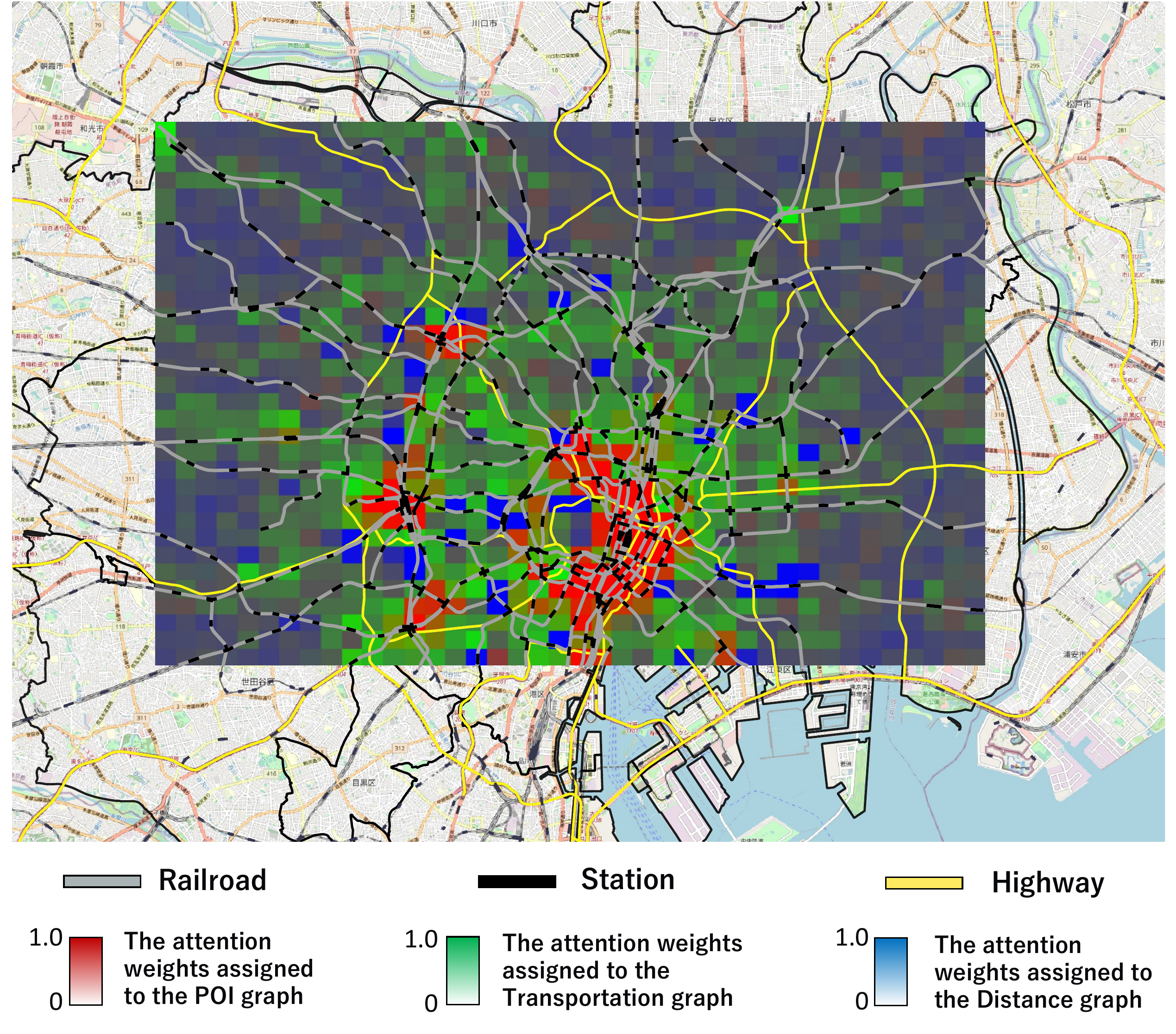}
    \caption{The attention weight with the real-world map. We show the weights for prediction at 10:00 a.m. on 2020/3/19 (Thursday).}
  \label{fig:attention_map}
\end{figure}

Further, we investigate how urban environmental information contributes to future population prediction at each region by analyzing the attention weights.
First, we depict the attention weight on a specific business day's morning with a map-based background in Figure \ref{fig:attention_map}. 

In the figure, we add some urban components such as railroads and highways on the map to visually understand the relationship between attention weights and urban features.
To keep the consistency with the model architecture depicted in 
Figure
\ref{fig:architecture},
we represent the weight value assigned to the distance graph in blue, and values assigned to the POI graph and the transportation graph in red and green, respectively. 
Note that when a particular color stands out in a specific area, it indicates that the model judges that corresponding urban environmental information is essential to predict the population there.
We depict attention weights in Figure \ref{fig:attention_map} that achieved the best performance in MAE and WAPE at $10$-times evaluation trials.

From Figure \ref{fig:attention_map}, we can see that the center of Tokyo tends to be highlighted in red or green color, which indicates that the POI graph and the transportation graph highly contribute to population prediction in those areas. Furthermore, the three spots highlighted in red, which are at the left side of the figure, are Ikebukuro, Shinjuku, and Shibuya, respectively. 
These places are known as the three major shopping and entertainment hubs of Tokyo. Given these characteristics, it is understandable that these areas are highlighted in red, indicating that the POI graph is essential to predict the future population.
Moreover, we can see that green highlighted regions (i.e., the regions where the transportation graph's contribution is significant) distribute similarly with the railroads.
Hence, we can be confident that the attention weights overall capture the real-world environment's characteristics.




\begin{figure}[h]
  \begin{tabular}{c}
    \begin{minipage}[t]{0.95\hsize}
      \centering
      \includegraphics[width=8.5cm]{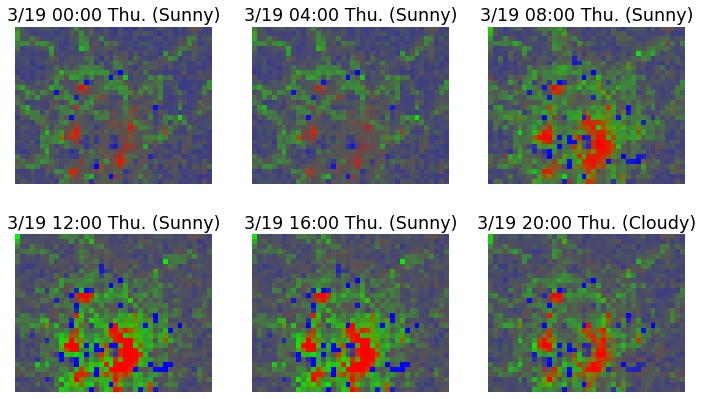}
      \subcaption{The attention weight on a business day with sunny weather}
      \label{fig:attention_normal}
    \end{minipage} \\
    \\
    \begin{minipage}[t]{0.95\hsize}
      \centering
      \includegraphics[width=8.5cm]{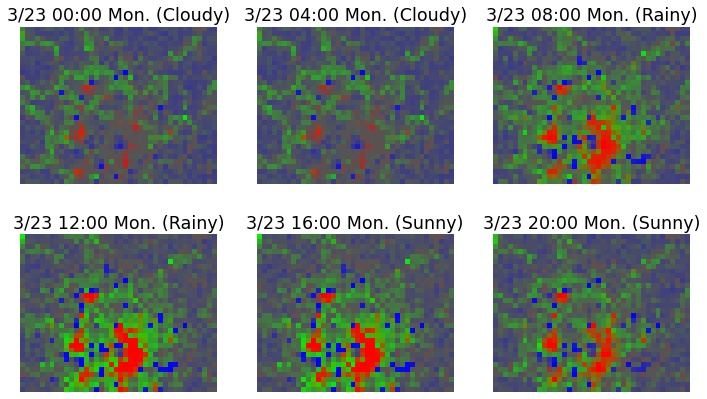}
      \subcaption{The attention weight on a business day with rainy weather}
      \label{fig:attention_rain}
    \end{minipage} \\
    \\
    \begin{minipage}[h]{0.95\hsize}
      \centering
      \includegraphics[width=8.5cm]{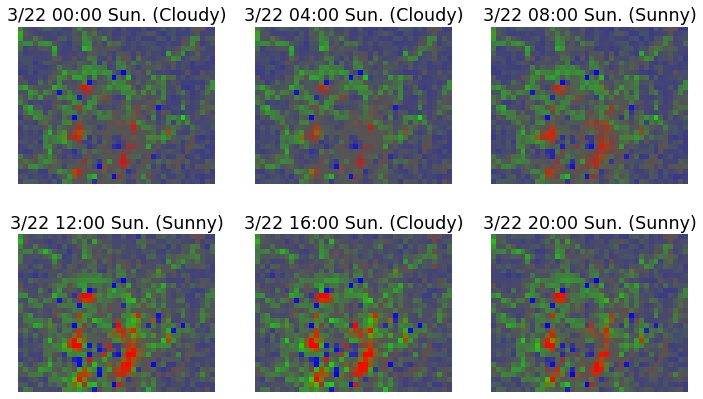}
      \subcaption{The attention weight on a holiday with sunny weather}
      \label{fig:attention_sun}
    \end{minipage} 
  \end{tabular}
  \caption{The visualized attention weights in various contextual situations. Blue indicates the weights assigned to the distance graph for prediction, and red and green indicate the weights assigned to the POI graph and the transportation graph, respectively.}
\end{figure}

We further investigate how the importance of the multiple kinds of urban environmental information will change in different contextual situations by visualizing the attention weights in various conditions.

Figure \ref{fig:attention_normal} shows the attention weights on a business day in sunny weather with multiple time intervals.
We can observe that red and green colors tend to become distinguished from the morning to the evening compared to midnight. 
Considering the business hours of the facilities and railroads, we can understand that this behavior of the attention module is reasonable.
Figure \ref{fig:attention_rain} shows the attention weights on a business day in rainy weather.
We notice no remarkable difference in the values of attention weights between Figure \ref{fig:attention_normal} and Figure \ref{fig:attention_rain} despite different weather conditions.
Therefore, we can understand that the model pays little attention to weather situations when it fuses multiple calculated values.
We can assume that one of the reasons for this result is seasonality.
We used population data collected from January to the end of March in Tokyo. In general, there is little precipitation in Tokyo in these seasons; thus, the effect of the weather situation might be pretty limited.
Besides, compared to other count-based data such as the number of taxi requests, the population figures may generally be little affected by the weather condition. 
Figure \ref{fig:attention_sun} shows the attention weights on a holiday in sunny weather.
We can observe that overall the weight assigned to the POI graph and the transportation graph is low compare to Figure \ref{fig:attention_normal}.
Specifically, this characteristic is remarkable in the center of Tokyo. 
We can conclude that this is because only a few people commute to the office on holiday, so the importance of the transportation graph for prediction is low compared to a business day.

From these observations, 
we can conclude that the model learns a valuable way to leverage urban environmental information by considering the contextual situations.
Surprisingly, the model automatically learns all of these trends without any prior knowledge about how to use urban environmental information.


\section{Conclusions and Future Work}
In this paper, we propose a novel method called ACMV-GCNs that predicts the short-term future population among a city by captures spatial correlations from multiple views based on urban environmental information.
The model efficiently fuses three predicted values (i.e., distance-based, POI-based, and transportation-based) utilizing the contextual-based attention module.
We investigate the performance of the proposed method on statistics population count data and demonstrate that it outperforms baseline methods.
By investigating the attention weights in multiple situations, 
we confirm that the model leverages urban environmental information for future population prediction in a very understandable manner.
Our future work will be investigating other kinds of urban environmental information that affect future populations and applying our model to other datasets.


\begin{acks}
  The authors would like to thank DOCOMO InsightMarketing, INC for their support of this project.
\end{acks}

\bibliographystyle{ACM-Reference-Format}
\bibliography{references}


\end{document}